\pdfoutput=1

\documentclass[11pt]{article}
\usepackage[dvipsnames]{xcolor}
\usepackage[final]{acl}

\usepackage{times}
\usepackage{latexsym}
\usepackage{amsmath}
\usepackage{enumitem}
\usepackage[T1]{fontenc}

\usepackage[utf8]{inputenc}

\usepackage{microtype}

\usepackage{inconsolata}

\usepackage{graphicx}
\usepackage{multirow}
\usepackage{booktabs}

\usepackage{amsmath}

\usepackage{algorithm}
\usepackage{algorithmic}
\usepackage{color}
\usepackage{xspace}
\usepackage{amssymb}

\newcommand{\model}[0]{\textsc{AutoDetect}\xspace}

\title{\textsc{AutoDetect}: Towards a Unified Framework for Automated Weakness Detection in Large Language Models}

\author{
Jiale Cheng$^{1,2}$\thanks{\ \ Equal contributions.} , Yida Lu$^{1,2}$\footnotemark[1] , Xiaotao Gu$^2$ , Pei Ke$^1$ , Xiao Liu$^{2,3}$ , \\ \textbf{ Yuxiao Dong$^3$ , Hongning Wang$^1$ , Jie Tang$^3$ , Minlie Huang$^1$}\thanks{\ \ Corresponding author.}\\
$^1$The Conversational Artificial Intelligence (CoAI) Group, Tsinghua University \\$^2$Zhipu AI\\
$^3$The Knowledge Engineering Group (KEG), Tsinghua University\\
\small{\texttt{{ \{chengjl23, lu-yd20\}@mails.tsinghua.edu.cn,}}} \small{\texttt{aihuang@tsinghua.edu.cn}}
\\
}

\newcommand{\hide}[1]{}

\newcommand{\blanksymbolfootnote}[1]{%
  \renewcommand{\thefootnote}{}
  \footnote{#1}%
  \setcounter{footnote}{0} 
  \renewcommand{\thefootnote}{\arabic{footnote}}
}

\begin{document}
\maketitle

\blanksymbolfootnote{$^2$Work done when JC and YL interned at Zhipu AI.}

\begin{abstract}
    
Although Large Language Models (LLMs) are becoming increasingly powerful, they still exhibit significant but subtle weaknesses, such as mistakes in instruction-following or coding tasks.
As these unexpected errors could lead to severe consequences in practical deployments, it is crucial to investigate the limitations within LLMs systematically.
Traditional benchmarking approaches cannot thoroughly pinpoint specific model deficiencies, while manual inspections are costly and not scalable. 
In this paper, we introduce a unified framework, \model, to automatically expose weaknesses in LLMs across various tasks. 
Inspired by the educational assessment process that measures students' learning outcomes, \model consists of three LLM-powered agents: Examiner, Questioner, and Assessor.
The collaboration among these three agents is designed to realize comprehensive and in-depth weakness identification. 
Our framework demonstrates significant success in uncovering flaws, with an identification success rate exceeding 30\% in prominent models such as ChatGPT and Claude.
More importantly, these identified weaknesses can guide specific model improvements, proving more effective than untargeted data augmentation methods like Self-Instruct. Our approach has led to substantial enhancements in popular LLMs, including the Llama series and Mistral-7b, boosting their performance by over 10\% across several benchmarks.
Code and data are publicly available at \url{https://github.com/thu-coai/AutoDetect}.
\end{abstract}



\section{Introduction}
    



\begin{figure}[t]
    \centering
    \includegraphics[width=\linewidth]{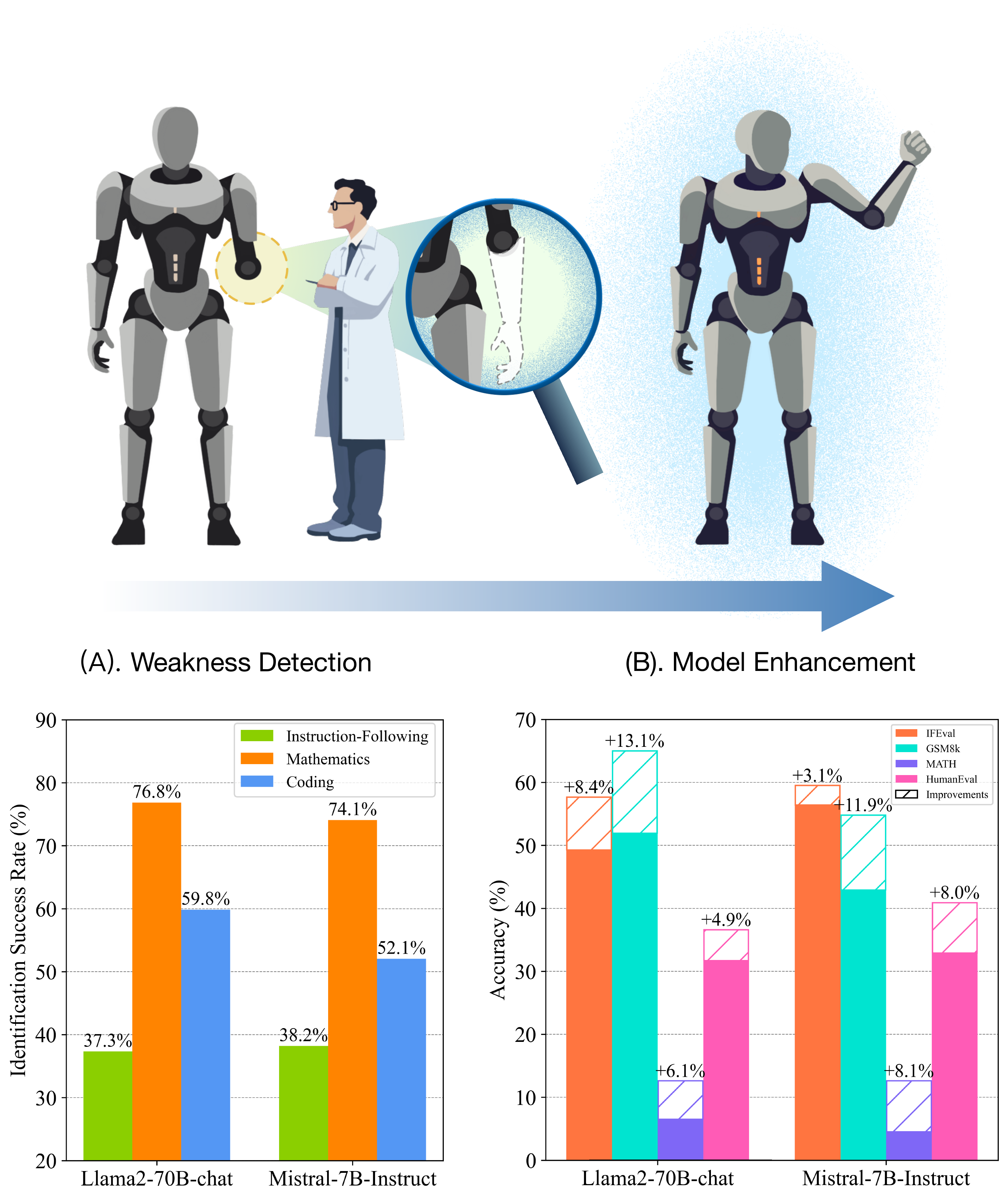}
    \caption{Effective weakness discovery can well guide model enhancement. \model can achieve high identification success rates in the instruction-following, mathematics, and coding tasks (A). Moreover, leveraging this data can further improve LLMs (B). 
    }
    \label{fig:weakness}
    \vspace{-5mm}
\end{figure}

The developments in Large Language Models (LLMs) are phenomenal: such models have demonstrated excellent performance in diverse tasks \cite{brown2020language, zeng2022glm, chowdhery2023palm, touvron2023llama, glm2024chatglm}.
After elaborate alignment \cite{ouyang2022training, cheng2023black, ji2024aligner}, LLMs can achieve human-level performance in real-world applications \cite{ChatGPT, Claude}.
Nevertheless, these models are prone to making unexpected mistakes \cite{ouyang2022training, bubeck2023sparks}.  
For instance, while LLMs are skilled at complex algorithm problems, they may struggle with basic coding concepts (\S\ref{para: limitations within LLMs}).
These unexpected mistakes can result in unforeseen consequences like system failures and significant safety issues \cite{ruan2023identifying}.
Consequently, systematically identifying and addressing these weaknesses is essential to enhancing the performance and trustworthiness of LLMs.


\begin{figure*}[t]
    \centering
    \includegraphics[width=0.95\linewidth]{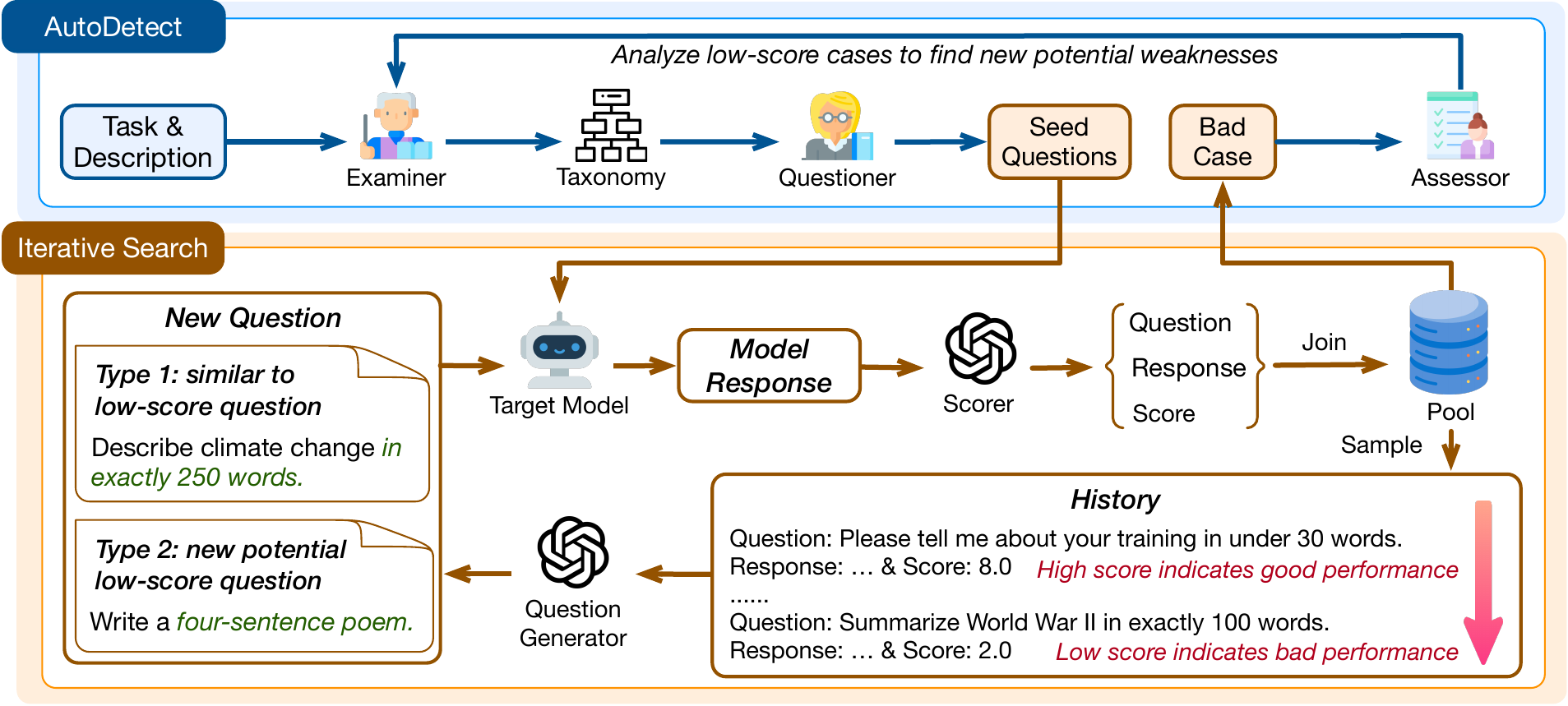}
    \caption{Our framework comprises two cycles, with the circulation consisting of the Examiner, Questioner, and Assessor, providing a comprehensive and tailored testing framework. Meanwhile, iterative search enables the adjustment of question difficulty for the target model, effectively identifying weaknesses.}
    \label{fig:framework}
    \vspace{-5mm}
\end{figure*}


However, the journey to reveal these weaknesses is challenging. 
Manual examinations rely on human experts, which are too labor-intensive and costly to scale. 
Automated methods, meanwhile, typically employ either static \cite{cobbe2021training, srivastava2022beyond, liu2023alignbench, gui2024logicgame} and dynamic \cite{bai2024benchmarking, wang2024benchmark} benchmarks. 
Unfortunately, benchmarks are intentionally structured to assess and rank a series of models, 
not to identify weaknesses inherent to individual models.
More specifically, benchmarks are designed to be model-agnostic without counting on specific model responses and thus inept at identifying individualized weaknesses.
Moreover, benchmarks suffer from infrequent updates, data leakage \cite{yang2023rethinking, wei2023skywork}, and leaderboard swamping \cite{guo2023evaluating}, which further limit their utility for thorough model-specific weakness assessment.

In this paper, we introduce a pioneering unified framework, \model, aiming to systematically and automatically expose potential weaknesses within LLMs across a variety of tasks.
In our framework, illustrated in Figure \ref{fig:framework}, we adopt a methodology analogous to educational assessment systems, comprising creating comprehensive questions to evaluate students and reviewing their responses to identify individualized weaknesses.
\model involves the development of a holistic testing system to assess and challenge student abilities. Moreover, this system is not static but constantly optimized and adapted to specific model performance, providing a tailored and effective weakness discovery.
Specifically, our framework integrates three specialized roles implemented by LLM-based agents:

\begin{itemize}[leftmargin=1.5em,itemsep=0pt,parsep=0.2em,topsep=0.1em,partopsep=0.0em]
    \item \textbf{Examiner} is tasked with building a comprehensive taxonomy featuring diverse test points and dynamically optimizing the framework based on the target model's performance in order to provide a refined and tailored framework for identifying potential weaknesses.
    
    \item \textbf{Questioner} is responsible for creating challenging questions according to each test point. 
    Through iterative explorations, this agent continually hypothesizes about the model's weaknesses, effectively adapting the generation of questions as new deficiencies emerge.
    
    \item \textbf{Assessor} needs to analyze the target model's responses and speculate on potential issues to be incorporated into the testing system, which is crucial to tailored assessments.
\end{itemize}
The collaboration among the Examiner, Questioner, and Assessor fosters an extensive and effective assessment process.
By learning from these weaknesses, \model further facilitates model improvements (Figure \ref{fig:weakness}).



Through extensive experiments, we demonstrate that \model is able to perform effective weakness exposure across a diverse range of tasks, including instruction-following, mathematical reasoning, and coding, achieving an impressive identification success rate of over 50\% in multiple strong LLMs and even more than 30\% in GPT-3.5-turbo and Claude-3-sonnet. 
Furthermore, our weakness identification process can effectively guide model enhancement.
Notably, by integrating about 1,000 samples derived from \model to fine-tune popular open-source models like Mistral and Llama series, we have achieved over 10\% improvements across several benchmarks, showing the benefits of learning from targeted weakness detection.


Our contribution can be summarized as follows:
\begin{itemize}[leftmargin=1.5em,itemsep=0pt,parsep=0.2em,topsep=0.1em,partopsep=0.0em]
    \item To the best of our knowledge, we are the first to systematically explore weakness identification in LLMs on multiple generic tasks, including instruction-following, mathematical reasoning, and coding, offering a unified framework for automatic weakness detection.
    \item \model has demonstrated exceptional adaptability and effectiveness, with a success rate of over 50\% in uncovering deficiencies across multiple models and tasks.
    \item \model facilitates significant model improvements. Leveraging the data derived from the weakness detection process, we can effectively enhance model performance, yielding over 10\% improvements on several tasks.
\end{itemize}

\section{Related Work}
    
\subsection{Evaluation Benchmarks}

Numerous benchmarks \cite{hendrycks2020measuring, cobbe2021training, chen2021evaluating, zhou2023instruction, liu2023alignbench, gui2024logicgame} are designed to evaluate various capabilities of LLMs, as well as some dynamic benchmarks \cite{zhu2023dyval, bai2024benchmarking, wang2024benchmark}.
However, the fundamental purpose of benchmarks is to compare a range of models and accurately rank them instead of identifying specific model defects. 
As a result, they are designed to be model-agnostic and cannot provide a thorough discovery of particular model flaws.
Additionally, static benchmarks often suffer from issues like data leakage \cite{yang2023rethinking, wei2023skywork} and leaderboard swamping \cite{guo2023evaluating}, while dynamic benchmarks often have trouble in coverage, and the methods used to construct these dynamic benchmarks usually lack universality. These limitations suggest that relying solely on benchmarks makes it challenging to thoroughly uncover model flaws, thus failing to offer practical guidance for further improvements.


%


\subsection{Red Teaming}
Due to the limitations of automated methods mentioned above, an essential way to effectively expose weaknesses within LLMs is manual checkup, which is similar to red teaming, an important strategy in the safety domain \cite{deng2023towards, ji2023ai, sun2023safety, ji2024beavertails} to identify safety issues of AI systems.
Early studies largely counted on manual efforts to create red teaming queries \cite{dinan2019build, xu2020recipes}.
However, manual red teaming is hampered by its high costs and inherent lack of diversity, limiting its scalability.
Recently, the use of language models for automated red team attacks has been proposed \cite{perez2022red} and widely adopted \cite{ganguli2022red, zhang2022constructing, chao2023jailbreaking}. Nevertheless, the application of automated weakness detection to general-purpose tasks remains underexplored. 
In this work, we introduce a unified framework for identifying model flaws beyond safety issues.
We have successfully implemented this framework in various tasks, including instruction-following, mathematical reasoning, and coding, demonstrating its impressive effectiveness and broad applicability.


\section{Method}

\subsection{Problem Definition}
Our primary objective is to develop a unified framework, aiming to automatically and systematically identify potential weaknesses in LLMs across generic tasks.
For a given task with its description, denoted as $(T, D)$, the weakness identification process can be represented as follows:
\begin{equation}
    \mathbb{W} = \model (T, D)
\end{equation}
where $\mathbb{W}$ stands for the set of problems that the target model fails to address accurately.
We consider these failures as model weaknesses, which are evaluated by a strong LLM judge.

\subsection{\model Framework}
The overall framework of our method is shown in Figure \ref{fig:framework}. \model is designed to comprehensively assess language model capabilities through a specialized circular search strategy, encompassing three distinct roles: \textit{Examiner}, \textit{Questioner}, and \textit{Assessor}. Each role is critical, leveraging the strengths of LLM-powered agents in a collaborative manner to explore and expose weaknesses in target models. 

As shown in Algorithm \ref{algo: pipeline}, the process begins with the Examiner, who is tasked with developing a detailed taxonomy $\mathbb{C}$ based on the given task and its description $(T, D)$. This taxonomy is crucial as it organizes the task into manageable, focused categories $(c_1, \cdots, c_n)$, each including several knowledge points $(k_1, \cdots, k_m)$, which guide the subsequent assessments. The structured decomposition is essential for thorough evaluation and is represented as:
\begin{equation}
    \mathbb{C} = \mathbf{Examiner}(T, D)
\end{equation}
Following taxonomy creation, the Questioner takes over, generating a seed set of questions $\mathbb{S}$ and initiating an iterative search process to craft questions $\mathbb{Q}$ that probe weaknesses at each knowledge point. The iterative process allows for adaptive questioning strategies that progressively increase in complexity, ensuring a depth of testing tailored to each model's capabilities. This can be formalized as:
\begin{equation}
    \mathbb{Q} = \mathbf{Questioner}(\mathbb{H})
\end{equation}
Here, $\mathbb{H}$ denotes the search history, starting from $\mathbb{S}$, facilitating a dynamic exploration of model weaknesses.
Importantly, the role of the Assessor is integral to refining the evaluation process to become thorough and model-specific. As the assessment progresses, the Assessor critically analyzes instances where the target model underperforms (indicated by low scores), identifying new potential weaknesses, $k_{new}$, expressed as:
\begin{equation}
    k_{new} = \mathbf{Assessor}(\mathbb{H}_{low})
\end{equation}
This insight leads to the Examiner's dynamic refinement of the taxonomy, ensuring that our framework stays relevant and effective at discovering new deficiencies. 
The cyclical interaction among the Examiner, Questioner, and Assessor realizes a continuous improvement loop, making our testing framework not only comprehensive but also sensitive to the evolving capabilities of different LLMs. Detailed descriptions of tasks and prompts in our framework are provided in Appendix \ref{appendix: task description} and Appendix \ref{appendix: detailed prompt}.

\begin{table*}[!t]
    \centering
    \resizebox{\linewidth}{!}{
        \begin{tabular}{l|ccc|ccc|ccc|c}
        \toprule
        \multirow{2}{*}{\textbf{Model}} & \multicolumn{3}{c|}{\textbf{Instruction following ISR (\%) $\downarrow$}} & \multicolumn{3}{c|}{\textbf{Mathematics ISR (\%) $\downarrow$}} & \multicolumn{3}{c|}{\textbf{Coding ISR (\%) $\downarrow$}} & \multirow{2}{*}{\textbf{Average ISR (\%) $\downarrow$}} \\ 
         & \textbf{Format} & \hspace{0.4em}\textbf{General} & \hspace{-0.4em}\textbf{Overall} & \textbf{Geometry} & \textbf{Analysis} & \textbf{Overall} & \textbf{DS.} & \hspace{0.5em}\textbf{MA.} & \textbf{Overall} &  \\ \midrule
        \multicolumn{11}{c}{\textit{Open-source Large Language Models}} \\ \midrule
        Llama2-7b-Chat & \color{BrickRed}\textbf{\large{55.3}} & \hspace{0.4em}\color{BrickRed}\textbf{\large{37.8}} & \hspace{-0.4em}\color{BrickRed}\textbf{\large{43.3}} & \color{BrickRed}\textbf{\large{89.8}} & \color{BrickRed}\textbf{\large{93.3}} & \color{BrickRed}\textbf{\large{88.8}} & \color{BrickRed}\textbf{\large{83.3}} & \hspace{0.5em}\color{BrickRed}\textbf{\large{81.1}} & \color{BrickRed}\textbf{\large{74.8}} & \color{BrickRed}\textbf{\large{69.0}} \\
        Llama2-13b-Chat & \large{52.4} & \hspace{0.4em}\large{35.0} & \hspace{-0.4em}\large{39.7} & \large{88.6} & \large{88.9} & \large{86.1} & \large{78.9} & \hspace{0.5em}\large{73.3} & \large{67.5} & \large{64.4} \\
        Llama2-70b-Chat & \large{51.3} & \hspace{0.4em}\large{34.9} & \hspace{-0.4em}\large{37.3} & \large{74.1} & \large{81.1} & \large{76.9} & \large{72.2} & \hspace{0.5em}\large{72.2} & \large{59.8} & \large{58.0} \\
        Mistral-7b-Instruct & \large{52.9} & \hspace{0.4em}\large{32.5} & \hspace{-0.4em}\large{38.2} & \large{77.8} & \large{71.1} & \large{74.1} & \large{66.7} & \hspace{0.5em}\large{56.7} & \large{52.1} & \large{54.8} \\
        Llama3-8b-Instruct & \large{42.1} & \hspace{0.4em}\large{19.4} & \hspace{-0.4em}\large{27.6} & \large{61.1} & \large{68.9} & \large{60.9} & \large{45.6} & \hspace{0.5em}\large{50.0} & \large{41.9} & \large{43.5} \\
        Llama3-70b-Instruct & {\color{ForestGreen} \textbf{\large{18.5}}} & \hspace{0.4em}{\color{ForestGreen} \textbf{\large{4.3}}} & \hspace{-0.4em}{\color{ForestGreen} \textbf{\large{10.2}}} & \large{41.9} & \large{30.0} & \large{38.7} & {\color{ForestGreen} \textbf{\large{15.6}}} & \hspace{0.5em}{\color{ForestGreen} \textbf{\large{16.7}}} & {\color{ForestGreen} \textbf{\large{15.7}}} & {\color{ForestGreen} \textbf{\large{21.5}}} \\ \midrule
        \multicolumn{11}{c}{\textit{Closed-source Large Language Models}} \\ \midrule
        GPT-3.5-turbo & \large{35.1} & \hspace{0.4em}\large{21.7} & \hspace{-0.4em}\large{25.5} & \large{56.3} & \large{35.6} & \large{50.2} & \large{40.0} & \hspace{0.5em}\large{30.0} & \large{32.5} & \large{36.1} \\
        Claude-3-sonnet & \large{29.3} & \hspace{0.4em}\large{12.8} & \hspace{-0.4em}\large{19.2} & \large{45.6} & \large{42.2} & \large{43.8} & \large{37.3} & \hspace{0.5em}\large{32.0} & \large{29.9} & \large{31.0} \\ 
        Mistral Large & \large{32.1} & \hspace{0.4em}\large{13.9} & \hspace{-0.4em}\large{20.3} & \large{41.5} & \large{30.0} & \large{33.9} & \large{38.9} & \hspace{0.5em}\large{33.3} & \large{26.4} & \large{26.8} \\ 
        GLM-4-Air & \large{32.0} & \hspace{0.4em}\large{9.2} & \hspace{-0.4em}\large{17.8} & {\color{ForestGreen} \textbf{\large{33.7}}} & {\color{ForestGreen} \textbf{\large{26.7}}} & {\color{ForestGreen} \textbf{\large{33.4}}} & \large{32.2} & \hspace{0.5em}\large{45.6} & \large{28.7} & \large{26.7} \\ \bottomrule
        \end{tabular}
    }
    \caption{ISR on multiple LLMs across the Instruction-following, Mathematics, and Coding tasks. We showcase the overall result and select two subtasks with the highest ISR. In each column, the highest ISR is highlighted in {\color{BrickRed} \textbf{red}} and the lowest in {\color{ForestGreen} \textbf{green}}. DS. denotes Data Structure and MA. refers to Mathematics and Algorithms.}
    \label{tab:mainres}
    \vspace{-5mm}
\end{table*}

\subsection{Iterative Search}
As the collaboration among the three roles guarantees the coverage and model-specificity of our framework, another crucial problem is how to effectively identify questions where the target model underperforms. 
Therefore, we leverage the strong exploration and evaluation capabilities of LLMs \cite{yang2023large, ke2023critiquellm, zheng2024judging} to develop an iterative search process.
Specifically, we first generate five questions for each knowledge point to create a seed set. The performance of the target model on this set is evaluated using a reference-based scoring method for reliability \cite{zheng2024judging}, where the reference responses are provided by GPT-4.
Subsequently, we rank historical samples by scores, with lower scores indicating poorer performance, to generate new questions that may expose model flaws. 
We then have the target model generate responses to the proposed question and score it, adding the result to our history collection. 
Through this iterative search process, we can effectively identify low-scoring questions, pinpointing specific weaknesses in the target model at particular knowledge points.

\subsection{Model Enhancement}
The ultimate goal of weakness discovery is to help models improve.
To validate that the identified weaknesses are non-trivial and can contribute to model enhancement, we further fine-tune the target model using the questions and reference answers obtained from the weakness detection process. Formally, the loss function is expressed as follows: 
\begin{equation}
    \mathcal{L}=-\frac{1}{N}\sum_{t=1}^N\text{log}P(R_t|Q,R_{<t}) 
\end{equation}
Here, $Q$ denotes the questions derived from the search, $R$ signifies the reference answers generated by GPT-4, and $N$ represents the length of $R$.

\section{Experiments}
    
To comprehensively demonstrate the superior performance of \model, we have conducted extensive experiments on diverse tasks: instruction-following, mathematics, and coding, including weakness detection (\S\ref{exp: weak detect}), model enhancement (\S\ref{exp: model improve}), comparisons with baseline methods (\S\ref{exp: baseline}), and iterative weakness recovery (\S\ref{exp: iterative}).
Implementation details can be found in Appendix \ref{appendix: Implementation}.

\subsection{Weakness Detection} \label{exp: weak detect}

\begin{table}[t]
\centering
\resizebox{0.9\linewidth}{!}{
\begin{tabular}{lcc}
\toprule
\textbf{Metric}                 & \textbf{Accuracy (\%)} & \textbf{Fleiss Kappa} \\ \midrule
Reasonableness      & 98.0         & 0.493              \\ \midrule
Agreement & 88.7        & 0.472              \\ \midrule
Correctness & 87.3        & 0.439              \\ \bottomrule
\end{tabular}}
\caption{Human evaluation results for the \model process. We evaluate for question reasonableness, agreement with GPT-4 evaluation, and the correctness of generated reference. Each Fleiss Kappa is greater than 0.4, indicating moderate agreement between annotators.}
\label{tab: human evaluation}
\vspace{-5mm}
\end{table}


We investigate three distinct tasks—instruction-following, mathematics, and coding—to demonstrate the generalization capabilities of \model.
The instruction-following task concentrates on providing the model with specific constraints, like formats and content. 
The mathematics task focuses on questions at a high school level, while the coding task focuses on Python in order to guarantee the correctness of the problems produced by GPT-4.

\begin{table*}[!t]
    \centering
    \setlength{\tabcolsep}{1mm}
    \renewcommand\arraystretch{1.15}
    \resizebox{\linewidth}{!}{
        \begin{tabular}{l|ll|ll|ll|ll|ll}
            \toprule
            \multicolumn{1}{l|}{\multirow{3}{*}{\textbf{Model}}} & \multicolumn{4}{c|}{\textbf{Instruction following}} & \multicolumn{4}{c|}{\textbf{Mathematics}} & \multicolumn{2}{c}{\textbf{Coding}} \\ 
            \multicolumn{1}{c|}{} & \multicolumn{2}{c}{\textbf{\hspace{-1em}IFEval-p}} & \multicolumn{2}{c|}{\hspace{-1em}\textbf{IFEval-i}} & \multicolumn{2}{c}{\hspace{-1em}\textbf{GSM8k}} & \multicolumn{2}{c|}{\hspace{-1em}\textbf{MATH}} & \multicolumn{2}{c}{\textbf{HumanEval}} \\ \cmidrule[0.5pt](l){2-5}\cmidrule[0.5pt](l){6-9}\cmidrule[0.5pt](l){10-11}  

            \multicolumn{1}{c|}{} & \multicolumn{1}{c}{ori.} & \multicolumn{1}{l}{\hspace{2.0em}ours} & \multicolumn{1}{c}{ori.} & \multicolumn{1}{l|}{\hspace{1.7em}ours} & \multicolumn{1}{c}{ori.} & \multicolumn{1}{l}{\hspace{2.0em}ours} & \multicolumn{1}{c}{ori.} & \multicolumn{1}{l|}{\hspace{1.6em}ours} & \multicolumn{1}{c}{ori.} & \multicolumn{1}{l}{\hspace{2.0em}ours} \\ \midrule
            
            \multicolumn{1}{l|}{Llama2-7b-Chat} & 32.3 & \hspace{1.0em}42.5 {\color{ForestGreen} \textsubscript{\textbf{(+10.2)}}} & 46.2 & \hspace{1.0em}54.7 {\color{ForestGreen} \textsubscript{\textbf{(+8.5)}}} & 18.9 & \hspace{1.0em}25.9 {\color{ForestGreen} \textsubscript{\textbf{(+7.0)}}} & 2.5  & \hspace{1.0em}4.7 {\color{ForestGreen} \textsubscript{\textbf{(+2.2)}}} & 13.4 & \hspace{1.0em}18.7 {\color{ForestGreen} \textsubscript{\textbf{(+5.3)}}} \\ 

            \multicolumn{1}{l|}{Llama2-13b-Chat} & 34.3 & \hspace{1.0em}43.3 {\color{ForestGreen} \textsubscript{\textbf{(+9.0)}}} & 45.8 & \hspace{1.0em}54.3 {\color{ForestGreen} \textsubscript{\textbf{(+8.5)}}} & 26.9 & \hspace{1.0em}33.7 {\color{ForestGreen} \textsubscript{\textbf{(+6.8)}}} & 3.9  & \hspace{1.0em}6.0 {\color{ForestGreen} \textsubscript{\textbf{(+2.1)}}} & 17.7 & \hspace{1.0em}24.4 {\color{ForestGreen} \textsubscript{\textbf{(+6.7)}}} \\ 

            \multicolumn{1}{l|}{Llama2-70b-Chat} & 44.2 & \hspace{1.0em}51.8 {\color{ForestGreen} \textsubscript{\textbf{(+7.6)}}} & 54.3 & \hspace{1.0em}63.5 {\color{ForestGreen} \textsubscript{\textbf{(+9.2)}}} & 51.9 & \hspace{1.0em}65.0 {\color{ForestGreen} \textsubscript{\textbf{(+13.1)}}} & 6.5 & \hspace{1.0em}12.6 {\color{ForestGreen} \textsubscript{\textbf{(+6.1)}}} & 31.7 & \hspace{1.0em}36.6 {\color{ForestGreen} \textsubscript{\textbf{(+4.9)}}} \\ 

            \multicolumn{1}{l|}{Mistral-7b-Instruct} & 51.2 & \hspace{1.0em}54.3 {\color{ForestGreen} \textsubscript{\textbf{(+3.1)}}} & 61.6 & \hspace{1.0em}64.7 {\color{ForestGreen} \textsubscript{\textbf{(+3.1)}}} & 42.9 & \hspace{1.0em}54.8 {\color{ForestGreen} \textsubscript{\textbf{(+11.9)}}} & 4.5  & \hspace{1.0em}12.6 {\color{ForestGreen} \textsubscript{\textbf{(+8.1)}}} & 32.9 & \hspace{1.0em}40.9 {\color{ForestGreen} \textsubscript{\textbf{(+8.0)}}} \\ 

            \multicolumn{1}{l|}{Llama3-8b-Instruct} & 70.1 & \hspace{1.0em}72.6 {\color{ForestGreen} \textsubscript{\textbf{(+2.5)}}} & 78.3 & \hspace{1.0em}79.7 {\color{ForestGreen} \textsubscript{\textbf{(+1.4)}}} & 75.4 & \hspace{1.0em}79.9 {\color{ForestGreen} \textsubscript{\textbf{(+4.5)}}} & 23.9  & \hspace{1.0em}27.1 {\color{ForestGreen} \textsubscript{\textbf{(+3.2)}}} & 55.5 & \hspace{1.0em}61.0 {\color{ForestGreen} \textsubscript{\textbf{(+5.5)}}} \\

            \multicolumn{1}{l|}{Llama3-70b-Instruct} & 76.9 & \hspace{1.0em}79.1 {\color{ForestGreen} \textsubscript{\textbf{(+2.2)}}} & 84.1 & \hspace{1.0em}85.5 {\color{ForestGreen} \textsubscript{\textbf{(+1.4)}}} & 92.2 & \hspace{1.0em}92.4 {\color{ForestGreen} \textsubscript{\textbf{(+0.2)}}} & 42.3 & \hspace{1.0em}46.9 {\color{ForestGreen} \textsubscript{\textbf{(+4.6)}}} & 79.3 & \hspace{1.0em}81.1 {\color{ForestGreen} \textsubscript{\textbf{(+1.8)}}} \\
            
            \bottomrule
        \end{tabular}
    }
    \caption{LLMs' performance on different benchmarks of three fundamental tasks before and after training with data derived from the identification process. We use prompt-level and instruction-level metrics with strict accuracy for the evaluation of the IFEval benchmark (denoted as IFEval-p and IFEval-i, respectively).}
    \label{tab:improvement}
    \vspace{-2mm}
\end{table*}

\paragraph{Evaluation Metrics}
In the iterative search process, we employ the scoring prompt from MT-bench \cite{zheng2024judging}, which achieves an 85\% agreement rate with human annotators. In our methodology, a score of three or below on a scale of ten indicates an error in the target model's response, as we additionally ask LLM not to score higher than three if the answer is wrong. We find an agreement rate over 88\% with humans when judging the correctness of model responses (Table \ref{tab: human evaluation}). Leveraging this, we define the identification success rate (ISR) as:
\begin{equation}
    ISR = \frac{Num_{<4}}{Num_{total}}
    \label{formula:asr}
\end{equation}
where $Num_{<4}$ denotes the count of responses rated below four, and $Num_{total}$ represents the total number of evaluations conducted.

\paragraph{Human Evaluation} 
To further validate the effectiveness of \model, we conduct a manual evaluation. We sample 150 pieces, with 50 from each task, across all LLMs. 
We hire three annotators to assess the following aspects:
\begin{itemize}[leftmargin=1.5em,itemsep=0pt,parsep=0.2em,topsep=0.1em,partopsep=0.0em]
    \item \textbf{Reasonableness}: Judging the logical coherence of the generated questions.
    \item \textbf{Agreement}: Determining if agree with the labels obtained using GPT-4 scoring, where a score no more than three represents an error.
    \item \textbf{Correctness}: Assessing the correctness of the reference answers.
\end{itemize}
The results (Table \ref{tab: human evaluation}) show that almost all questions generated by \model are considered reasonable, with over 87\% of the reference answers being correct. Moreover, there is high agreement (88.7\%) with the labels obtained based on GPT-4 scoring. The annotation document for the evaluation process can be found in Appendix \ref{appendix: annotation_doc}.


\begin{figure}[ht]
    \centering
    \includegraphics[width=0.95\linewidth]{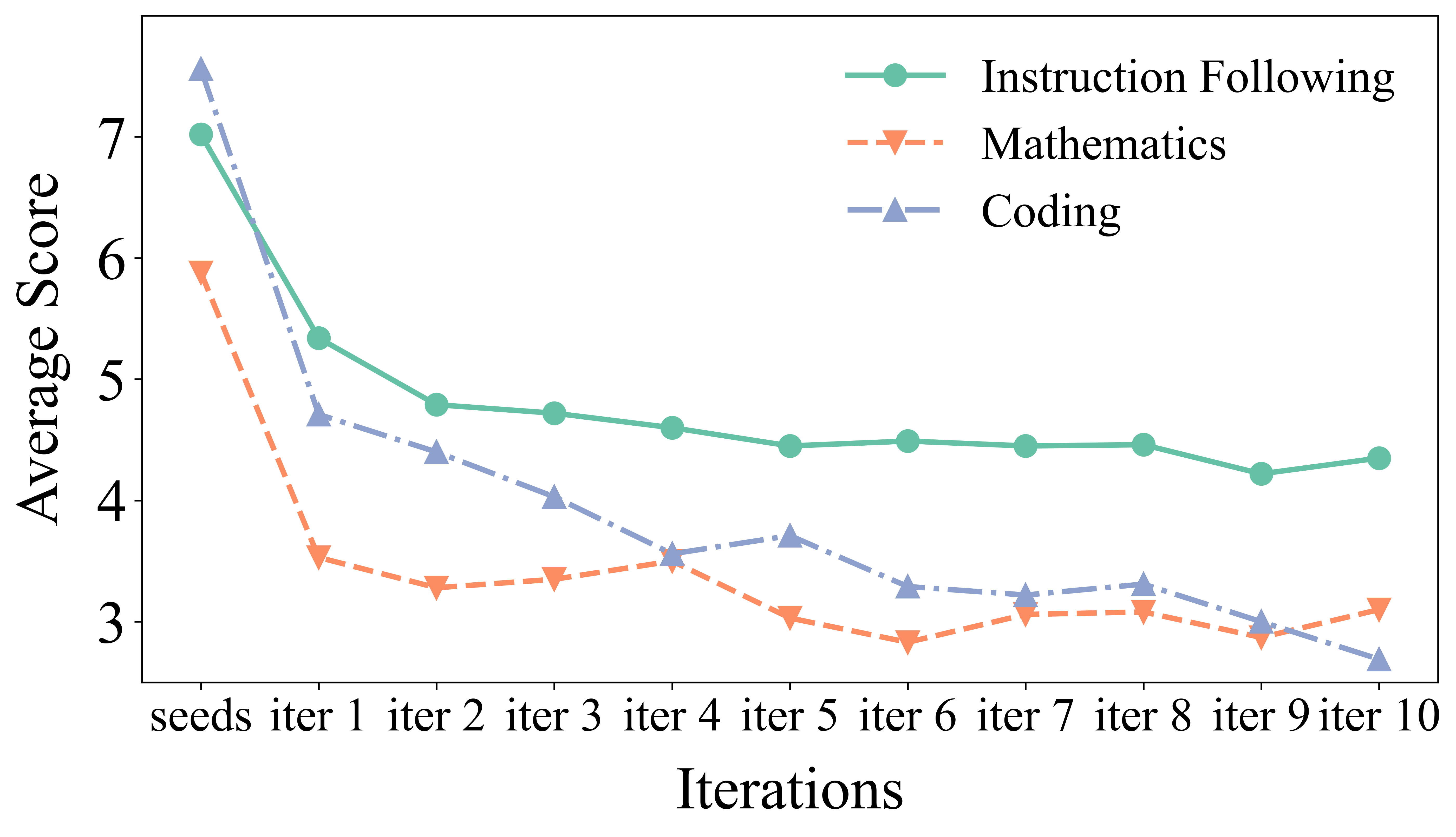}
    \caption{The change in the average score during the iterative search process for the three tasks.}
    \label{fig:score_change}
    \vspace{-3mm}
\end{figure}

\paragraph{Results}
As shown in Table \ref{tab:mainres}, we conduct flaw exploration across multiple models and achieve impressive ISR in various tasks, demonstrating the effectiveness of \model. 
Interestingly, the average score and ISR align well with the Chatbot Arena's model rankings \cite{chiang2024chatbot}, with a Spearman correlation of 0.95 between the two rankings, showing the potential of our approach as a dynamic benchmark.
As illustrated in Figure \ref{fig:score_change}, we present the average scores throughout the iterative search process.
The evident downward trend in scores highlights the significant role of the iterative method in uncovering model weaknesses.

\begin{table*}[t]
    \centering
    \resizebox{\linewidth}{!}{
        \begin{tabular}{l|cccc|cccc|ccc}
        \toprule
        \multirow{4}{*}{\textbf{Method}} & \multicolumn{4}{c|}{\textbf{Instruction Following}} & \multicolumn{4}{c|}{\textbf{Mathematics}} & \multicolumn{3}{c}{\textbf{Coding}} \\ \cmidrule(lr){2-5} \cmidrule(lr){6-9} \cmidrule(lr){10-12}
         & \multirow{2}{*}{\textbf{ISR (\%)} $\uparrow$} & \multicolumn{2}{c}{\textbf{Improvement} $\uparrow$} & \multirow{2}{*}{\textbf{BLEU-4} $\downarrow$} & \multirow{2}{*}{\textbf{ISR (\%)} $\uparrow$} & \multicolumn{2}{c}{\textbf{Improvement} $\uparrow$} & \multirow{2}{*}{\textbf{BLEU-4} $\downarrow$} & \multirow{2}{*}{\textbf{ISR (\%)} $\uparrow$} & \textbf{Improvement} $\uparrow$ & \multirow{2}{*}{\textbf{BLEU-4} $\downarrow$} \\ \cmidrule(lr){3-4} \cmidrule(lr){7-8} \cmidrule(lr){11-11}
         &  & \textbf{IFEval-p} & \textbf{IFEval-i} &  &  & \textbf{GSM8k} & \textbf{MATH} &  &  & \textbf{HumanEval} &  \\ \midrule
        Self-instruct & \large{20.4} & \large{35.7} & \large{47.4} & \underline{\large{0.40}} & \large{71.5} & \large{21.5} & \large{3.9} & \large{0.66} & \large{38.7} & \large{14.6} & \large{0.87} \\
        OPRO & \textbf{\large{72.9}} & \large{34.8} & \large{47.1} & \large{0.41} & \large{93.2} & \large{21.3} & \underline{\large{4.0}} & \underline{\large{0.48}} & \textbf{\large{95.1}} & \large{14.0} & \textbf{\large{0.38}} \\
        PAIR & \underline{\large{62.3}} & \underline{\large{37.2}} & \underline{\large{50.5}} & \large{0.45} & \underline{\large{95.2}} & \underline{\large{24.6}} & \large{3.0} & \large{0.62} & \large{83.3} & \underline{\large{15.2}} & \large{0.69} \\
        Ours & \large{56.8} & \textbf{\large{42.5}} & \textbf{\large{54.7}} & \textbf{\large{0.25}} & \textbf{\large{96.1}} & \textbf{\large{25.9}} & \textbf{\large{4.7}} & \textbf{\large{0.42}} & \underline{\large{92.4}} & \textbf{\large{18.7}} & \underline{\large{0.46}} \\ \bottomrule
        \end{tabular}
    }
    \caption{Results of comparison with baselines. The ISR of our method is the success rate of the iterative process. \textbf{Bold} indicates the best results and \underline{underline} means the second best.}
    \label{tab:baseline}
    \vspace{-5mm}
\end{table*}


\subsection{Model Enhancement} \label{exp: model improve}
To validate the identified flaws are meaningful and facilitate model enhancement, we fine-tune the models using data obtained during the \model process and evaluate them on popular benchmarks. Importantly, we do not use any data from the test sets.

\paragraph{Backbone Models}
The Llama series of models \cite{touvron2023llama2, llama3}, including the Llama2-chat with 7b, 13b, and 70b parameters, and Llama3-Instruct ranks the most popular. The Mistral-7b-Instruct \cite{jiang2023mistral} stands out as one of the best-performing models of its size.

\paragraph{Evaluation Benchmarks}
In our work, we evaluate the capability of instruction-following using the IFEval dataset \cite{zhou2023instruction}, which consists of 541 verifiable instructions. For mathematics, we choose the most popular benchmarks, GSM8k \cite{cobbe2021training} and MATH \cite{hendrycks2021math}.
In the coding task, we employ the widely-used HumanEval \cite{chen2021evaluating} for evaluation, which includes 164 carefully designed test cases created by human experts.

\begin{figure}[htbp]
    \centering
    \includegraphics[width=0.95\linewidth]{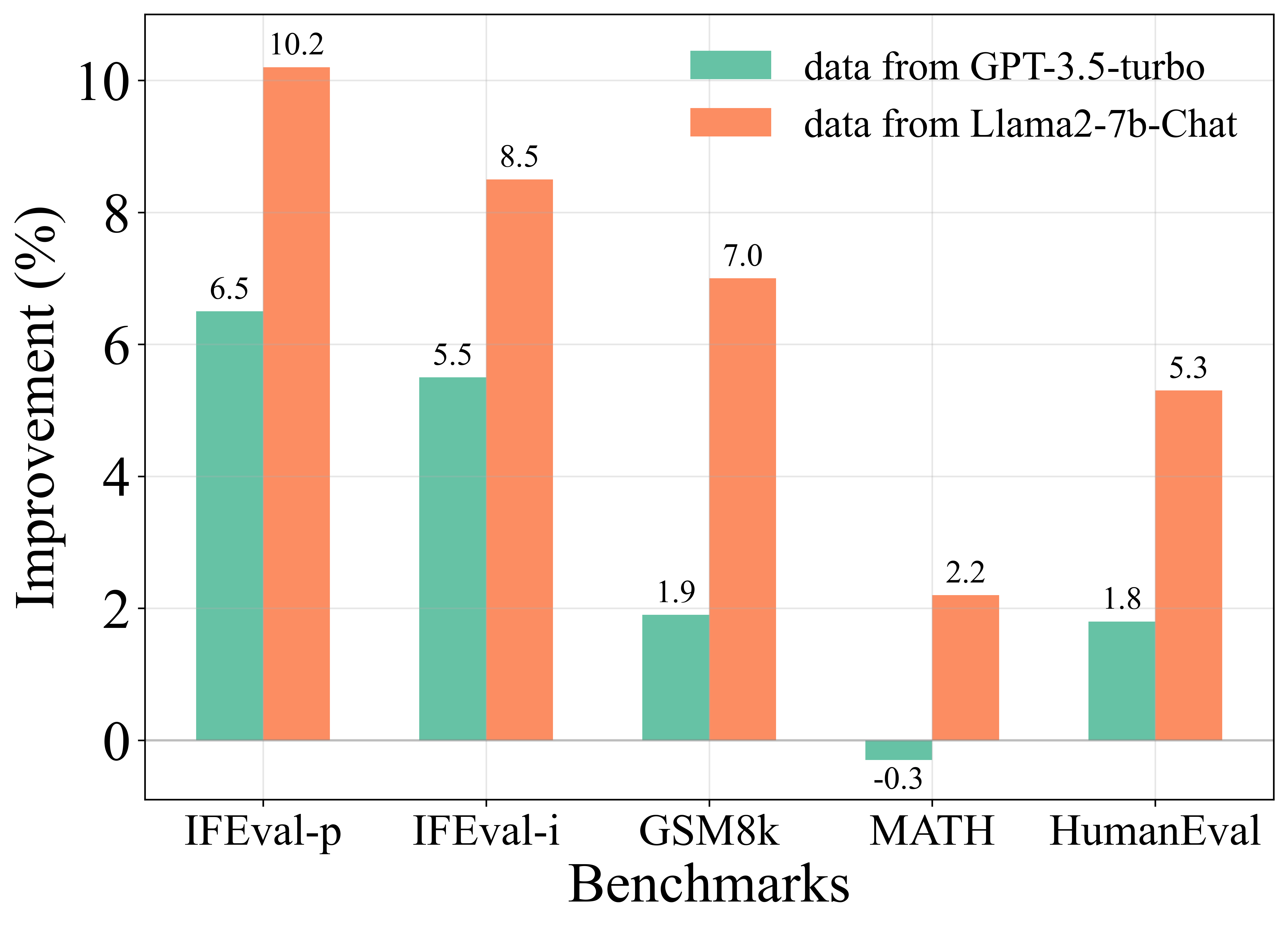}
    \caption{Improvement of Llama2-7b-Chat when training with identification data from GPT-3.5-turbo and itself.}
    \label{fig:comp35_improve}
    \vspace{-5mm}
\end{figure}

\paragraph{Results}
As shown in Table \ref{tab:improvement}, data from \model process enable us to significantly improve model performance. We achieve remarkable improvements across multiple models and tasks.
Moreover, the similar performance improvements, averaging over 6\% across test sets, for various sizes of Llama2 models confirm that our method remains effective as the model scales.
Furthermore, we investigate the impact of using assessment data from other models to boost the performance of the \texttt{llama2-chat-7b} model. The results, as shown in Figure \ref{fig:comp35_improve}, indicate that the effectiveness of using targeted assessment data is obviously superior to using \texttt{gpt-3.5-turbo}.
We also identified a distribution discrepancy in the identification data across various target models, which is detailed in Appendix \ref{appendix: distribution_discrepancy}. 
These findings highlight that targeted assessment can expose specific weaknesses in models, and addressing them leads to more significant improvements in model performance.

\subsection{Comparison with Baselines} \label{exp: baseline}
\paragraph{Baselines}
Self-Instruct \cite{wang2023self} is a widely-used method for data augmentation; OPRO \cite{yang2023large} applies iterative search optimization with LLMs; PAIR \cite{chao2023jailbreaking} is a popular method in the safety field for automatic jailbreak attacks and we transfer it to our tasks. We implement these baselines with their standard settings and maintain the same setting of training as \model to ensure a fair comparison.

\paragraph{Results}
As shown in Table \ref{tab:baseline} and Appendix \ref{appendix: cat_classification}, when compared to baselines, \model demonstrates superior performance in both identification success rate and diversity. 
Self-Instruct exhibits a low ISR and limited diversity. Meanwhile, OPRO and PAIR focus on exploiting specific weaknesses repeatedly, resulting in unbalanced problem distributions. While they can achieve high ISRs, they fail to provide a meaningful assessment across varied categories, limiting the utility for comprehensive weakness detection. In addition, PAIR is three times more costly than the other methods. 
Moreover, considering the improvements, \model outperforms others significantly, indicating that \model can comprehensively discover various weaknesses and provide more guidance in model enhancement.


\begin{table}[t]
    \centering
    \resizebox{0.8\linewidth}{!}{
        \begin{tabular}{l|ll}
        \toprule
        \textbf{Iteration} & \hspace{0.5em}\textbf{IFEval-p} & \hspace{0.5em}\textbf{IFEval-i} \\ \midrule
        Iter 0 (ori.) & \hspace{1em}34.3 & \hspace{1em}45.8 \\
        Iter 1 & \hspace{1em}43.3 \textcolor[RGB]{82,190,128}{\textsubscript{\textbf{(+9.0)}}} & \hspace{1em}54.3 \textcolor[RGB]{82,190,128}{\textsubscript{\textbf{(+8.5)}}} \\
        Iter 2 & \hspace{1em}45.4 \textcolor[RGB]{82,190,128}{\textsubscript{\textbf{(+2.1)}}} & \hspace{1em}57.0 \textcolor[RGB]{82,190,128}{\textsubscript{\textbf{(+2.7)}}} \\
        Iter 3 & \hspace{1em}47.1 \textcolor[RGB]{82,190,128}{\textsubscript{\textbf{(+1.7)}}} & \hspace{1em}58.2 \textcolor[RGB]{82,190,128}{\textsubscript{\textbf{(+1.2)}}} \\ \bottomrule
        \end{tabular}
    }
    \caption{Iterative improvement of Llama2-13b-Chat on IFEval benchmark.}
    \label{tab:iter_improve}
    \vspace{-5mm}
\end{table}

\begin{figure*}[ht]
    \centering
    \includegraphics[width=0.95\linewidth]{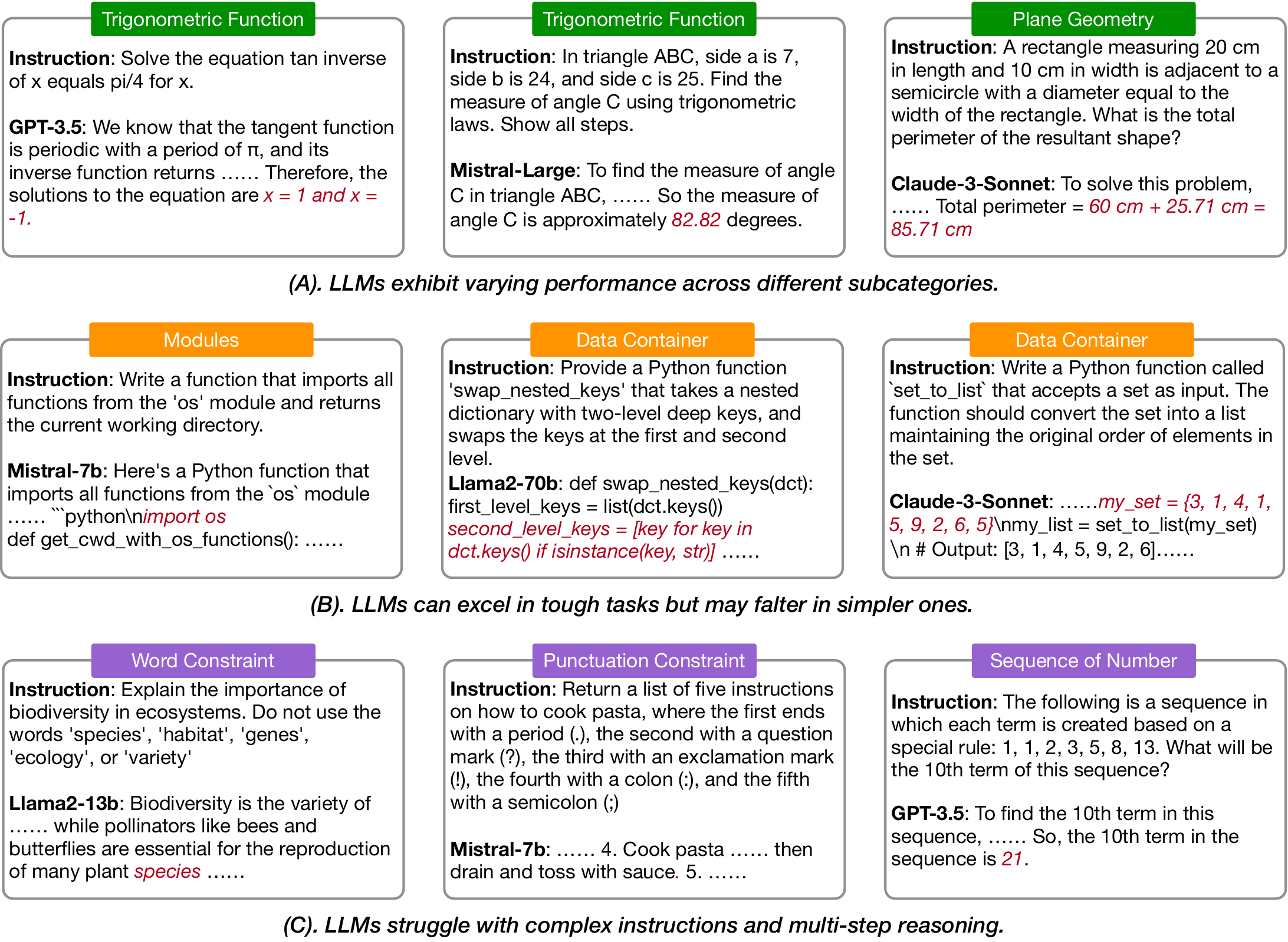}
    \caption{Some weaknesses within LLMs revealed by \model. We flag the wrong parts of the responses in red, and some responses are omitted due to space restrictions.}
    \label{fig: bad case}
    \vspace{-3mm}
\end{figure*}

\begin{figure*}[!ht]
    \centering
    \includegraphics[width=0.95\linewidth]{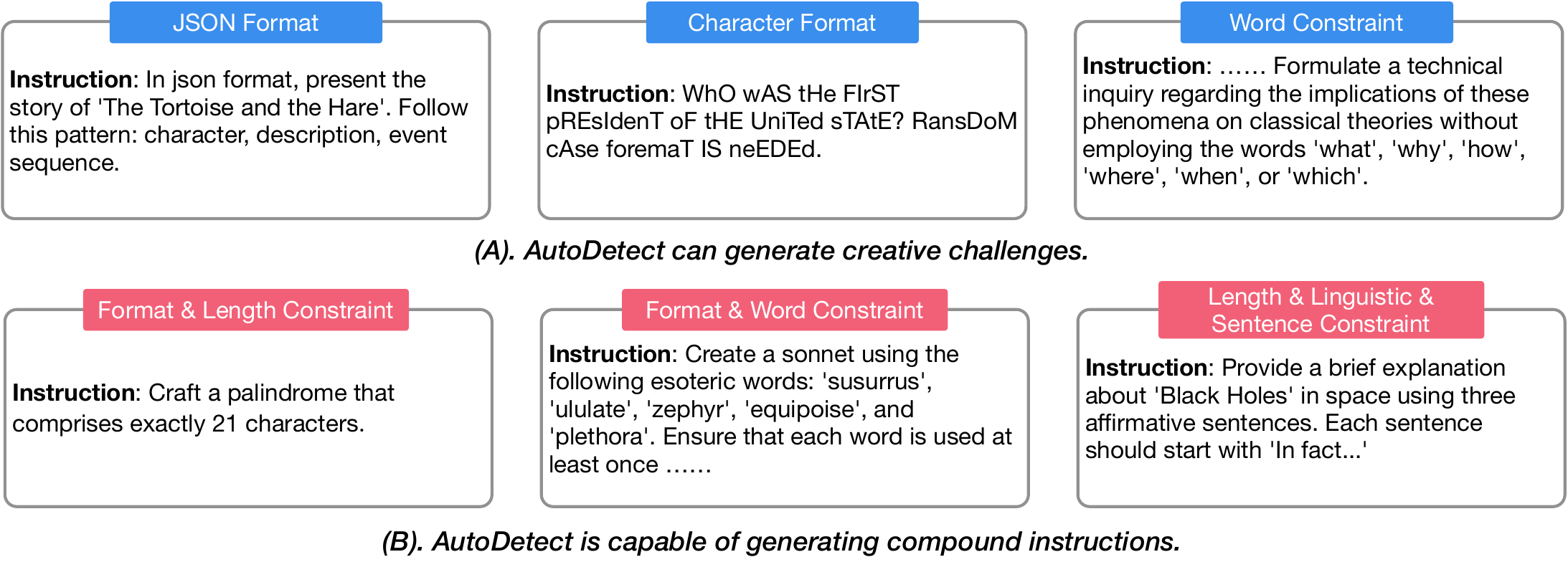}
    \caption{\model demonstrates some superior capabilities, such as generating creative and compound instructions. We omit some instructions due to space limits.}
    \label{fig: good case}
    \vspace{-3mm}
\end{figure*}

\vspace{-2mm}
\subsection{Iterative Weakness Recovery} \label{exp: iterative}
Since our framework can identify and help address the weaknesses of LLMs, a natural question arises: \textit{Can we iteratively improve the model's performance through \model?} We thus conduct the experiment on \texttt{llama2-13b-chat} in the instruction-following task. As shown in Table \ref{tab:iter_improve}, we observe that \model could consistently improve the model with three rounds of assessments. 
Furthermore, each iteration yields a non-trivial improvement, demonstrating the remarkable scalability of our approach. The improvements on categories during iterations are detailed in Appendix \ref{appendix: iterative_improvement}.

\section{Discussion}

With \model, we systematically identify potential weaknesses across various models.
Our comprehensive analysis reveals several notable findings, including the limitations within LLMs (\S\ref{para: limitations within LLMs}, shown in Figure \ref{fig: bad case}) and strengths of \model (\S\ref{para: superiority of AutoDetect}, shown in Figure \ref{fig: good case}), which may facilitate further research.

\vspace{-2mm}
\subsection{Limitations within LLMs} \label{para: limitations within LLMs}

\paragraph{LLMs exhibit varying performance across task subcategories.} 
In the mathematics task, models like \texttt{gpt-3.5-turbo}, \texttt{mistral-large} and \texttt{claude-3-sonnet} exhibit remarkable abilities, achieving over 90\% accuracy in benchmarks such as GSM8k. 
However, despite their strong performance on math word problems, we find these models struggle with simple geometric-related questions. For example, angle computation problems often lead to errors. 
This discrepancy in performance within the same domain demonstrates critical limitations in the comprehensiveness of benchmark-based weakness discovery.











\paragraph{LLMs can excel in tough tasks but may falter in simpler ones.}
Our findings indicate that LLMs can handle some difficult and complex challenges, like coding problems involving algorithms and data structures. However, these models show great misunderstandings and errors in simpler and even fundamental tasks. 
For instance, an unexpected error occurs when a model is instructed to import all functions from a specific module, and it merely imports the module itself. Similarly, when asked to write a function operating a set, it adds duplicate elements during the initialization of a set.
These issues suggest a critical flaw: \textit{LLMs' ability in complex tasks does not guarantee proficiency in simpler operations}, highlighting the necessity for further examinations into the boundaries of LLMs' capabilities to avoid potential risks.







\paragraph{LLMs struggle with complex instructions and multi-step reasoning.} 
LLMs continue to exhibit shortcomings when it comes to executing instructions with complete accuracy, especially those comprising several constraints or multi-step reasoning. They often omit parts of the instructions or make mistakes in later steps of multi-step tasks. This indicates their limited ability to perform in complex scenarios, which is essential for agent tasks.





\vspace{-2mm}
\subsection{Superiority of \model} \label{para: superiority of AutoDetect}


\paragraph{\model can generate creative challenges.}
When conducting automatic weakness identification, we find that \model could generate complex and unique questions that surpass typical human-written ones, especially for non-expert annotators. 
For example, telling a story in JSON format is a creative exercise, and human annotators are likely to be limited in their ability to think of such instructions.
This ability to generate diverse, challenging questions can be used to assess a model's advanced capabilities and further construct high-quality training data, thereby enhancing the model's performance.





\paragraph{\model is capable of generating compound instructions.}
Interestingly, we note the emergence of compound tasks in generated problems. In the instruction-following task, although we do not require the model to combine different constraints, we observe a few spontaneous combinations, both inter-category and intra-category types.
For instance, while LLMs perform well in translation tasks, their effectiveness diminishes when asked to translate into multiple languages simultaneously.




\section{Conclusion}


In this work, we introduce a unified framework, \model, for identifying weaknesses across various models and diverse tasks, including instruction-following, mathematical reasoning, and coding. Leveraging our method, we not only successfully uncover specific weaknesses but also effectively enhance the model performance using the data from the assessment process. Our results highlight the potential of using large language models to automatically detect and address model weaknesses on general tasks, helping us better understand the boundaries of model capabilities and paving the way for automatic LLM alignment.

\section*{Limitations}

Despite the strong capabilities of \model in identifying and addressing LLMs' weaknesses, showing the potential of leveraging AI to align AI, we want to discuss some known limitations, which need to be resolved through future research.


\paragraph{Enhancing the robustness of \model.} 
Even though the human evaluation results show most generated problems are reasonable, a small number of illogical questions, such as unsolvable math problems, may still occur. In addition, while our experiment shows that \model can stably discover model weaknesses with high ISR in repeated experiments (Appendix \ref{appendix: robustness}), the problems detected vary. This may need to be further validated with larger-scale weakness identifications.

\paragraph{Identifying weaknesses in more advanced models.}
Although \model can identify weaknesses in strong LLMs such as GPT-3.5 and Claude-3, the current framework—comprising three agents tasked with pinpointing weaknesses and generating data for model enhancement—heavily relies on the efficacy of a strong LLM. Consequently, the effectiveness of \model in detecting weaknesses is intrinsically limited by the agents' capabilities. When the target model's performance is on par with or surpasses that of the agents, uncovering its vulnerabilities becomes markedly more challenging. This challenge is exacerbated in self-evolution scenarios, where self-evaluation bias \cite{zheng2024judging, panickssery2024llm} can lead the model to overrate the quality of its own outputs. Innovative approaches are necessary to break these constraints and mitigate self-evaluation bias, which remains an underexplored area of research. We leave this for future work.


\section*{Ethical Considerations}
    In the weakness discovery process, \model generates test cases from scratch without using any existing datasets, and there are no license issues.
Our work focuses on generic tasks and does not involve security tasks, so there are no security concerns. 
\model is designed to discover models' weaknesses in general tasks, thus helping figure out the potential issues of LLMs and improving their robustness.
\model is designed to identify vulnerabilities in models across general tasks, allowing for the detection of potential issues within LLMs, ultimately aiding in their robustness enhancement.
In the human evaluation process, we hired three Chinese annotators, made the payment according to the regional standard, and informed the purpose of the experiment.

\section*{Acknowledgements}
This work was supported by the National Science Foundation for Distinguished Young Scholars (with No. 62125604). This work was also supported by the NSFC projects (Key project with No. 61936010). We would also like to thank Zhipu AI for sponsoring GPU computing and API costs consumed in this study. 

\bibliography{custom}

\appendix

\appendix

\section{\model Algorithm}
The overall process of \model is shown in Algorithm \ref{algo: pipeline}. 
\model primarily consists of two cycles and involves three roles: the Examiner, the Questioner, and the Assessor. 
The interaction among these LLM agents realizes a comprehensive and targeted assessment framework. Moreover, the iterative search conducted by the Questioner guarantees the effectiveness of weakness identification.

\begin{algorithm*}
\caption{\model Weakness Identification}
\begin{algorithmic}
\STATE $\mathbb{C} \leftarrow \mathbf{Examiner}(T, D) \quad \triangleright \text{Break down detailed taxonomy}$
\STATE $\mathbb{W} \leftarrow \emptyset \quad\quad\quad\quad\quad\quad\quad\quad \triangleright \text{The set of identified weaknesses} $
\FOR{each category $c \in \mathbb{C}$}
    \STATE $\mathbb{K} \leftarrow \text{get\_knowledge\_points}(c) \quad \triangleright \text{Initialize with knowledge points under this subcategory}$
    \WHILE{ $K \neq \emptyset $ }
        \STATE $\mathbb{S} \leftarrow \text{gen\_seed\_questions}(\text{pop}(\mathbb{K})) \quad \triangleright \text{Generate some seed questions of each knowledge point}$
        \STATE $\mathbb{H} \leftarrow \text{get\_resp\_score}(\mathbb{S}) \quad \triangleright \text{Get response from target model and score from scorer}$
        \FOR{Iterative search of $N$ turns} 
            \STATE $\mathbb{Q} = \mathbf{Questioner}(\mathbb{H}) \quad \triangleright \text{Generate a new challenging question}$
            \STATE $ \text{update}(\mathbb{H}, \mathbb{W}) \quad \triangleright \text{Generate responses and scores for new questions, then update } \mathbb{H} \text{ and } \mathbb{W} $
        \ENDFOR
        \STATE $k_{new} \leftarrow \mathbf{Assessor}(\mathbb{H}_{low}) \quad \triangleright \text{Analyze low-score cases to find new potential weaknesses}$
        \STATE $\mathbb{K} \leftarrow \mathbb{K} \cup k_{new}$
    \ENDWHILE
\ENDFOR
\RETURN $\mathbb{W}$
\end{algorithmic}
\label{algo: pipeline}
\end{algorithm*}

\section{Task Taxonomy and Description} \label{appendix: task description}
\begin{figure*}[!t]
    \includegraphics[width=\linewidth]{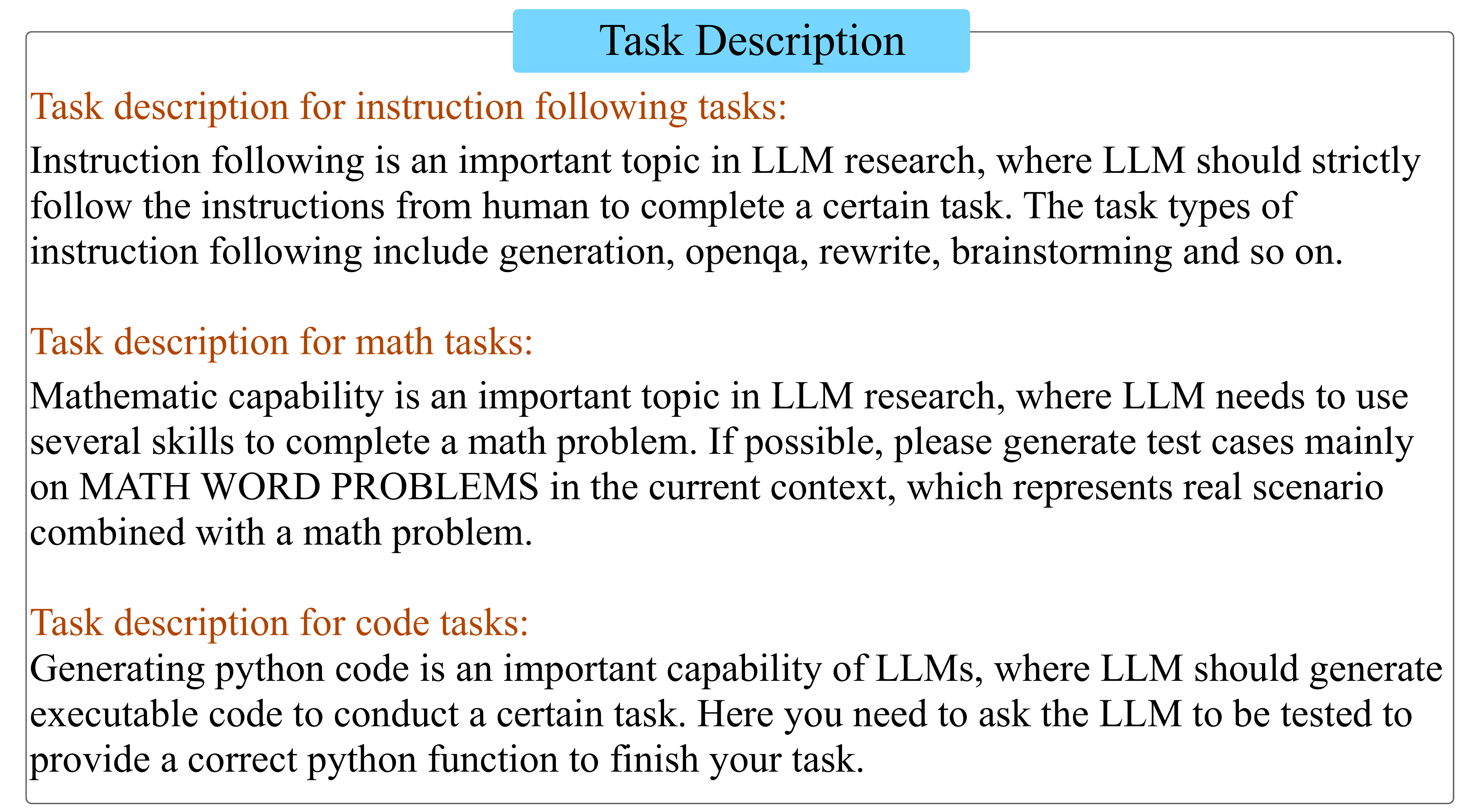}
    \caption{Description for the Instruction-following, Mathematics, and Coding tasks. }
    \label{fig:task_desc}
\end{figure*}

\begin{figure*}[t]
    \centering
    \includegraphics[width=\linewidth]{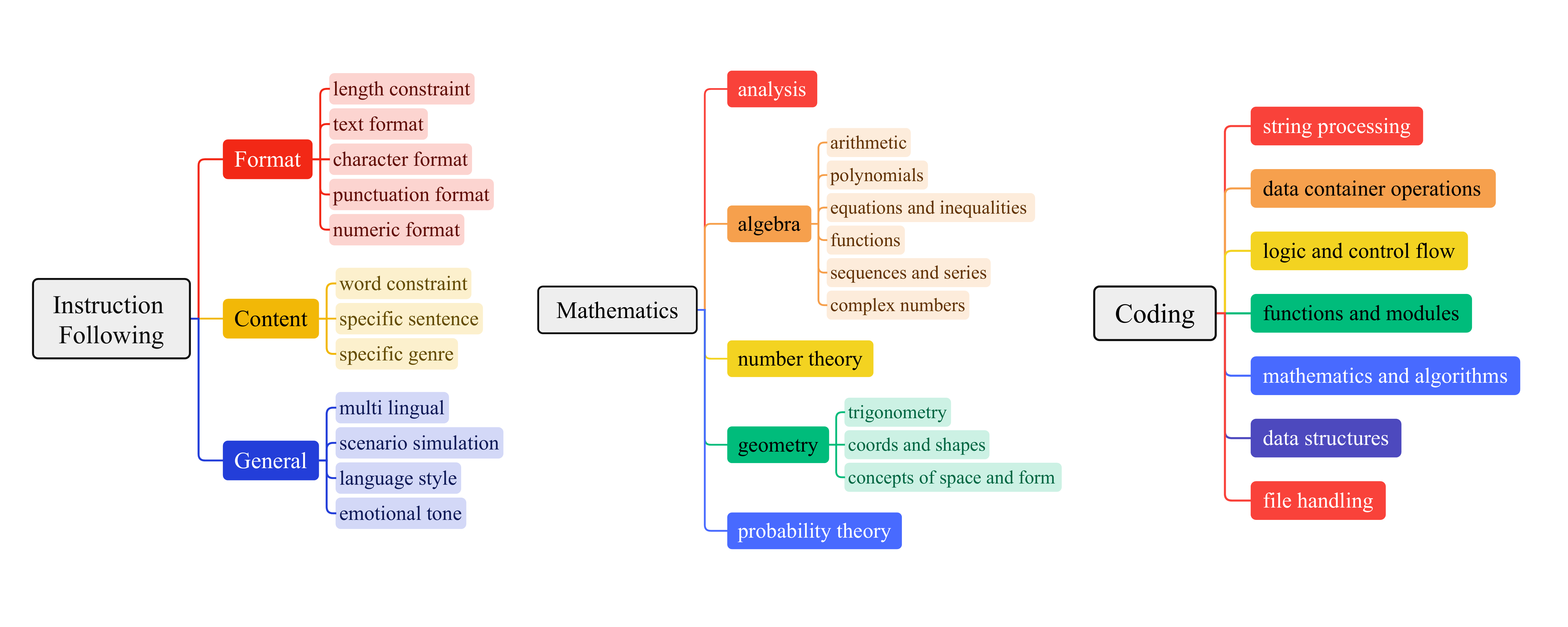}
    \caption{Detailed taxonomy used in our framework for the Instruction-following, Mathematics, and Coding tasks. }
    \label{fig:taxonomy}
\end{figure*}

For instruction-following, mathematics, and coding tasks, we show the descriptions and taxonomies in Figure \ref{fig:task_desc} and Figure \ref{fig:taxonomy}. The taxonomies are automatically generated by the Examiner with minimal human revision.

\section{Detailed Prompts} \label{appendix: detailed prompt}
\begin{figure*}[t]
    \centering
    \includegraphics[width=\linewidth]{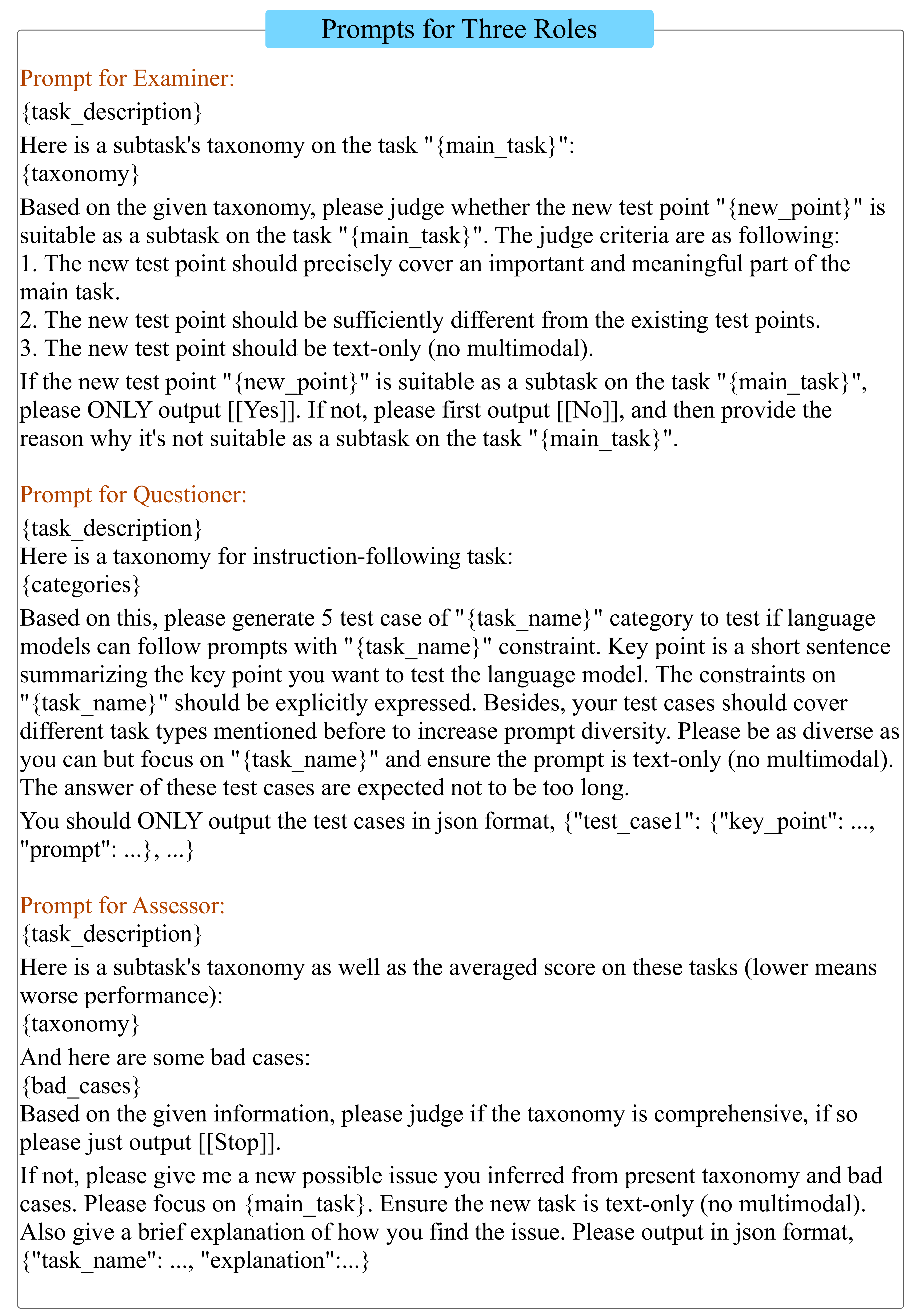}
    \caption{Prompts for the three roles: Examiner, Questioner, and Assessor in our framework. Words in the braces will be replaced with correlative content in real practice.}
    \label{fig:prompt_roles}
\end{figure*}

\begin{figure*}[ht]
    \centering
    \includegraphics[width=0.95\linewidth]{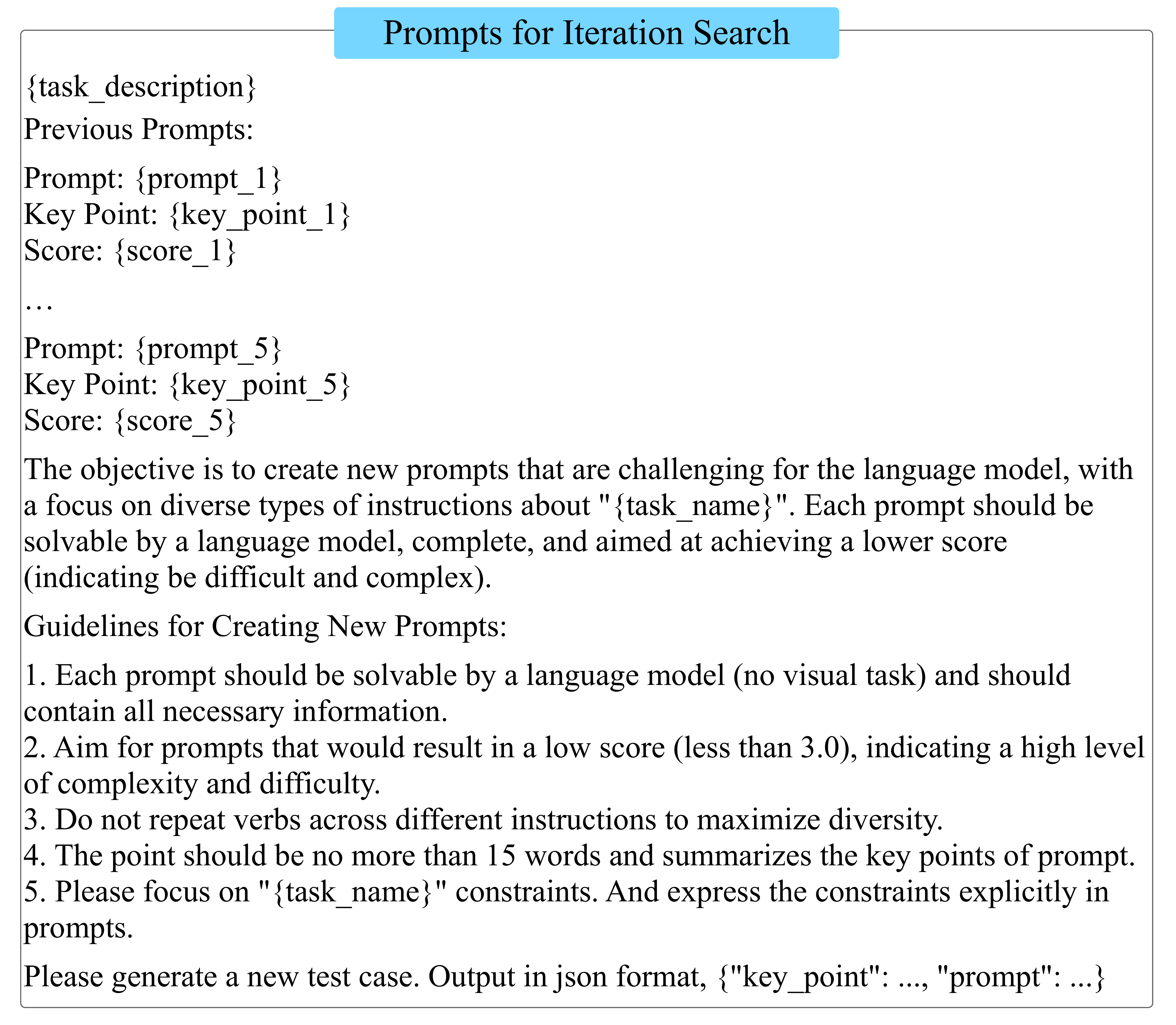}
    \caption{Prompt for the iteration search process in our framework.}
    \label{fig:prompt_iter}
\end{figure*}

We show all detailed prompts for each role in Figure \ref{fig:prompt_roles}, and the prompt used in the iterative search is shown in Figure \ref{fig:prompt_iter}.

\begin{table*}[!h]
    \centering
    \resizebox{0.9\linewidth}{!}{
        \begin{tabular}{l|c|c|c}
        \toprule
        \multicolumn{1}{c|}{\textbf{Llama2-7b-Chat}} & \textbf{Instruction Following ISR (\%)} & \textbf{Mathematics ISR (\%)} & \textbf{Coding ISR (\%)} \\ \midrule
        Repeat 1 & 43.3 & 88.8 & 74.8 \\
        Repeat 2 & 47.7 & 87.8 & 75.1 \\
        Repeat 3 & 45.5 & 87.7 & 79.2 \\ \bottomrule
        \end{tabular}
    }
    \caption{Repeated experiment for Llama2-7b-Chat on Instruction-Following, Mathematics and Coding task. }
    \label{tab:robustness}
\end{table*}
\begin{table*}[!t]
    \centering
    \resizebox{0.4\linewidth}{!}{
        \begin{tabular}{l|c|c}
        \toprule
        \textbf{Categories} & \textbf{Before} & \textbf{After} \\ \midrule
        Scenario Simulation (\%) & 64.4 & 32.2 \\
        Multi Lingual (\%) & 78.9 & 58.9 \\
        Word Constraint (\%) & 44.4 & 25.6 \\
        Specific Sentence (\%) & 35.6 & 25.6 \\
        Text Format (\%) & 65.6 & 57.8 \\
        Overall (\%) & 50.2 & 43.4 \\
        \bottomrule
        \end{tabular}
    }
    \caption{Top 5 categories and overall ISR changes on Llama2-13b-Chat before and after iterative improvement.}
    \label{tab:iter_improve_on_cat}
    \vspace{-5mm}
\end{table*}

\section{Implementation Details} \label{appendix: Implementation}


We employ the \texttt{gpt-4-1106-preview} for identifying weaknesses to ensure the performance of our framework. We limit each subcategory to a maximum of six knowledge points to balance coverage and cost. During the iterative search process, we utilize 5 seed questions and conduct a search over 10 steps to balance effectiveness and cost.
During the training phase, we choose LoRA fine-tuning \cite{hu2021lora}, as full-parameter fine-tuning significantly impacts performance due to limited data. We utilize the AdamW \cite{loshchilov2017decoupled} optimizer with $\beta_1$ set at 0.9 and $\beta_2$ at 0.999. To accelerate training, we adopt the Deepspeed Zero 2 strategy \cite{rasley2020deepspeed}. In addition, we train all models for 5 epochs with a batch size of 4. We use a learning rate of 2e-5, along with 0.1 warm-up steps and a linear decay schedule.
In the inference phase, we employ the vllm framework \cite{kwon2023efficient} to speed up output and employ greedy decoding. All the experiments are conducted on 8×80GB NVIDIA A100 GPUs.

\section{Baseline Category Classification} \label{appendix: cat_classification}
\begin{figure*}[ht]
    \centering
    \includegraphics[width=0.95\linewidth]{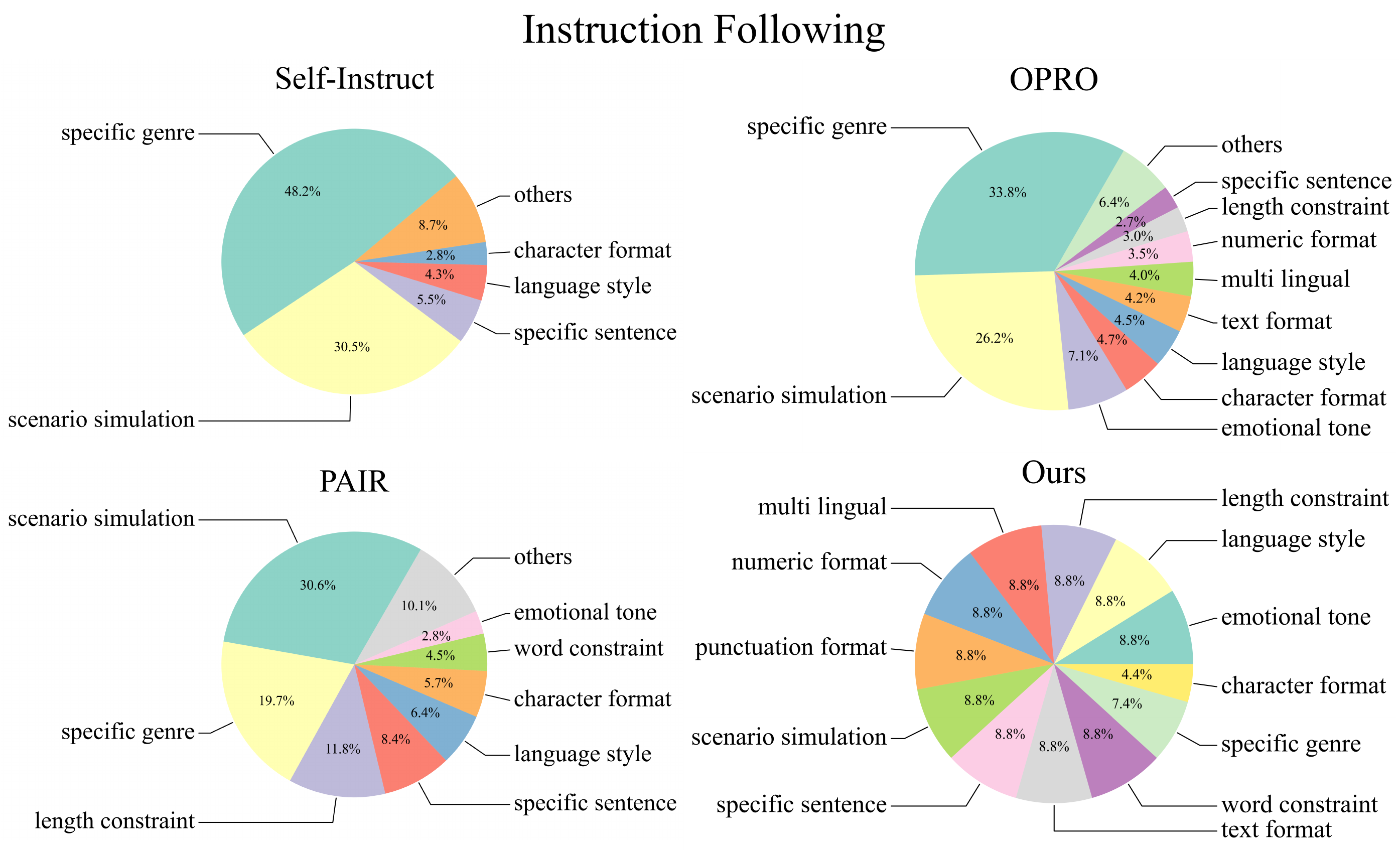}
    \caption{Classification of generated cases on three baselines and our method on Instruction Following task.}
    \label{fig:classification_pie_if}
\end{figure*}

\begin{figure*}[ht]
    \centering
    \includegraphics[width=\linewidth]{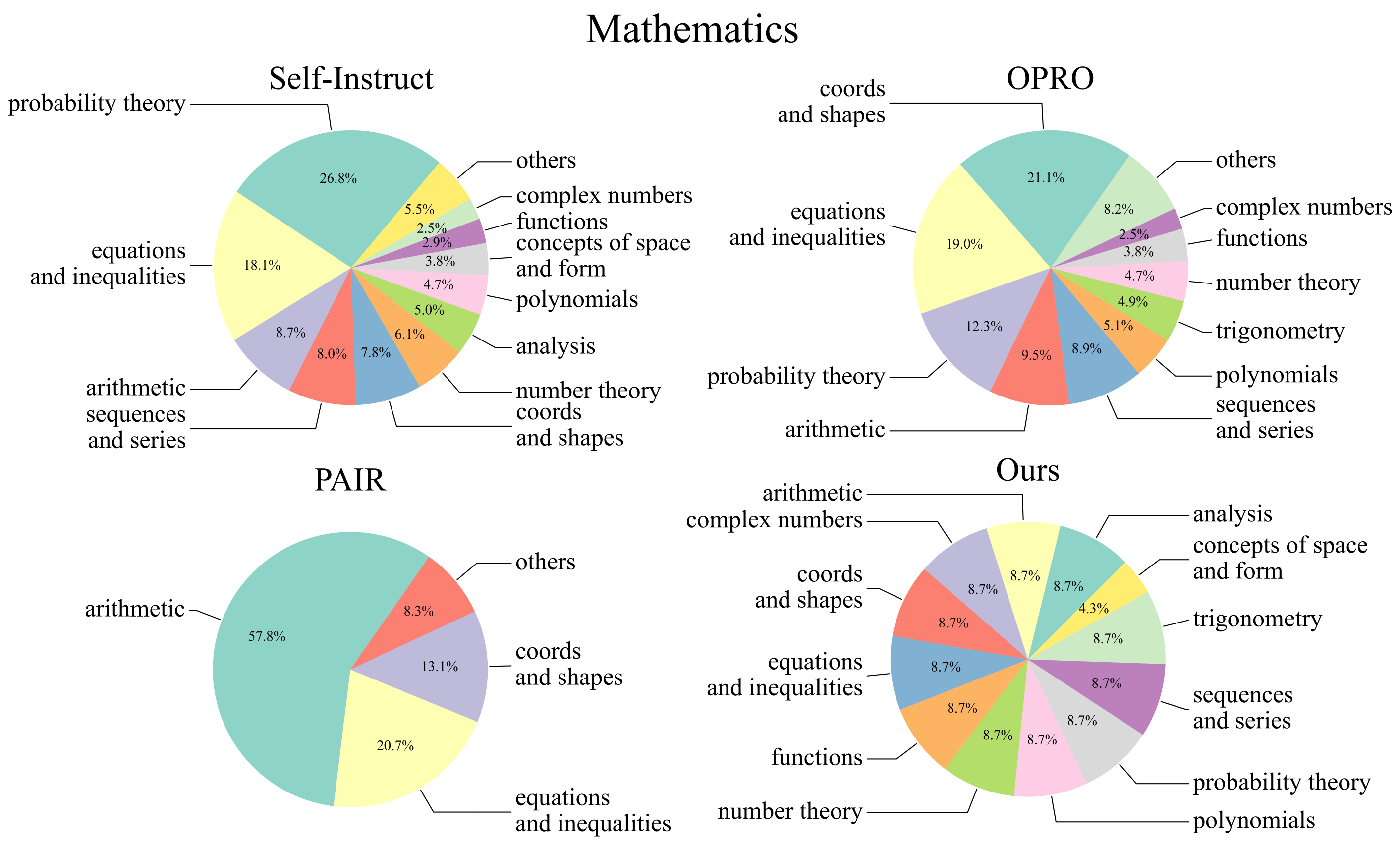}
    \caption{Classification of generated cases on three baselines and our method on Mathematics task.}
    \label{fig:classification_pie_math}
\end{figure*}

\begin{figure*}[ht]
    \centering
    \includegraphics[width=\linewidth]{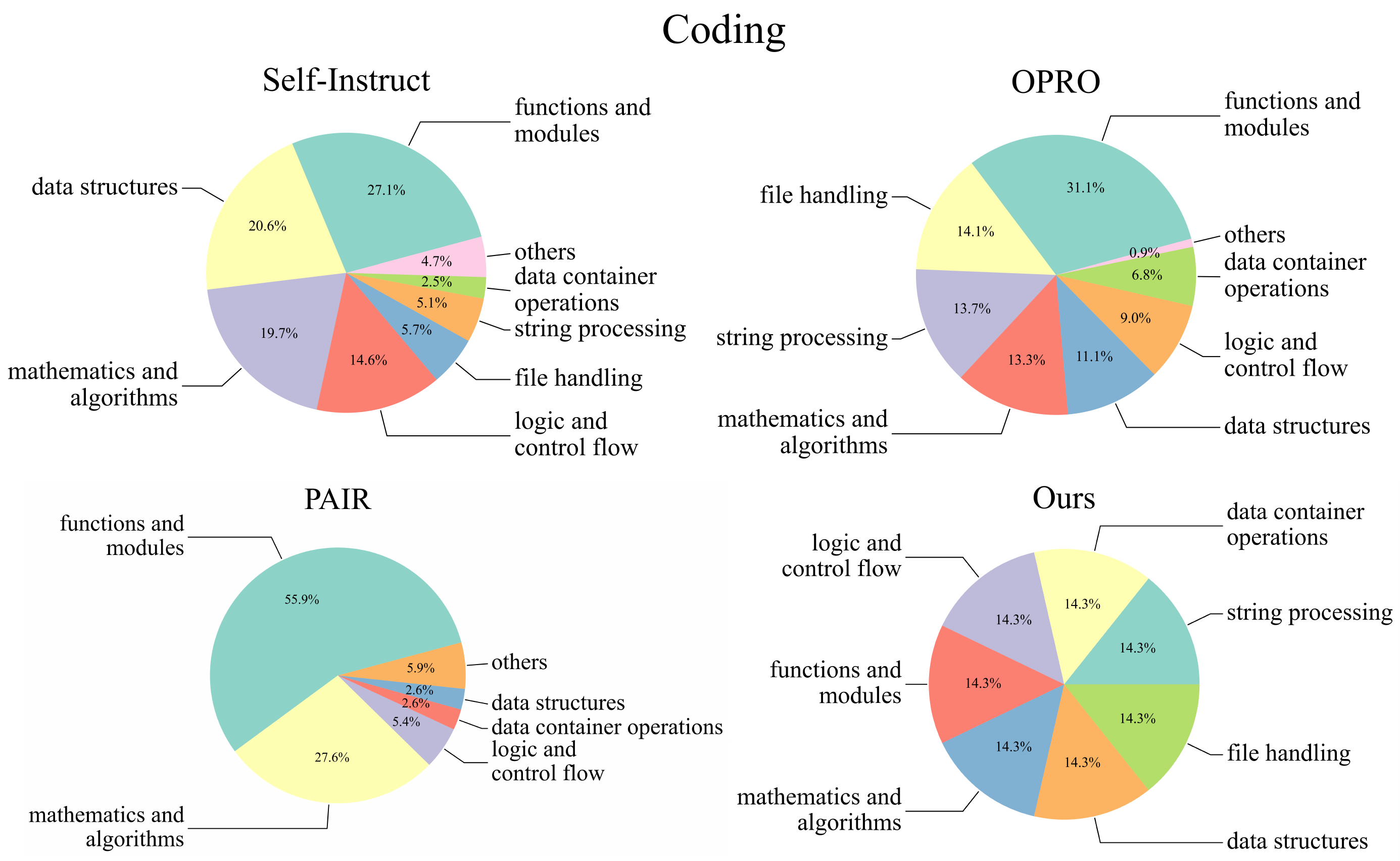}
    \caption{Classification of generated cases on three baselines and our method on Coding task.}
    \label{fig:classification_pie_code}
\end{figure*}

To further explore the diversity of generated cases on three baselines mentioned in \S\ref{exp: baseline}, we employ GPT-3.5-turbo for classification and map all the cases to our task taxonomy. We merge cases that don't belong to any category as well as categories that contain less than 2.5\% cases of the total into the category ``others''. The classification result is shown in Figure \ref{fig:classification_pie_if}, \ref{fig:classification_pie_math} and \ref{fig:classification_pie_code}. As our results exhibit, \model maintains a balanced distribution and a high diversity on all three tasks compared with these baselines.

\section{Robustness Experiment} \label{appendix: robustness}
To verify the robustness of our framework, we conduct repeated experiments on \texttt{llama2-7b-chat}. The result is shown in Table \ref{tab:robustness}. The slight fluctuation in the identification success rate across three tasks demonstrates the remarkable stability of our framework. 

\section{Annotation Document}
\label{appendix: annotation_doc}

\begin{figure*}[ht]
    \centering
    \includegraphics[width=\linewidth]{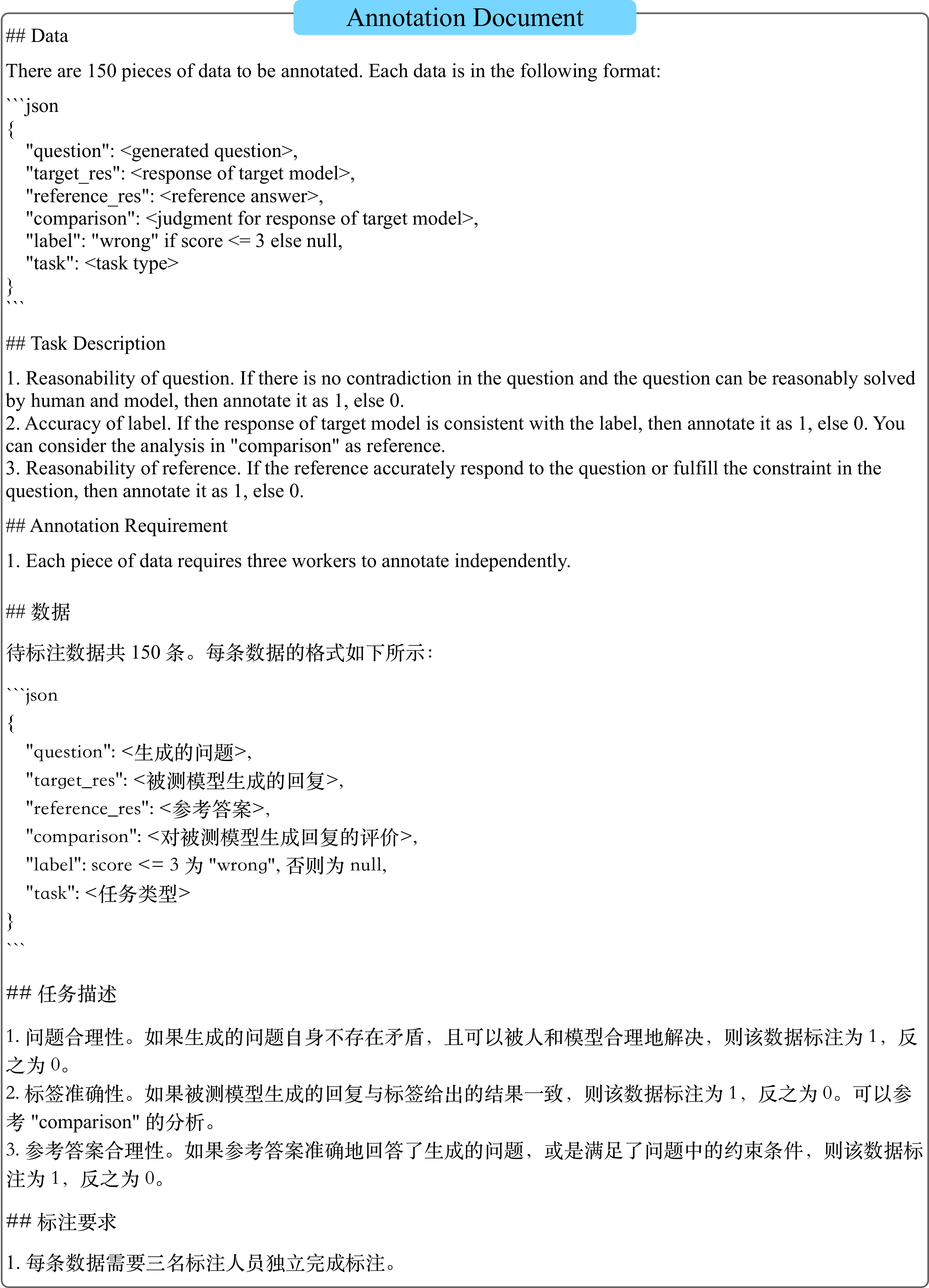}
    \caption{Annotation document for the human evaluation process in \S\ref{exp: weak detect}}
    \label{fig:annotation}
\end{figure*}

We provide the annotation document for the human evaluation process in \S\ref{exp: weak detect} in Figure \ref{fig:annotation}.

\section{Distribution Discrepancy in Identification Data}
\label{appendix: distribution_discrepancy}

We have observed a notable distribution discrepancy in the categories identified across different target models. For example, within the \textit{Scenario Simulation} subcategory of the instruction-following task, the Assessor identified new categories based on each model's specific weaknesses:
\begin{itemize}[leftmargin=1.5em,itemsep=0pt,parsep=0.2em,topsep=0.1em,partopsep=0.0em]
    \item \textbf{Llama2-7B-Chat}: Business Scenarios, Humorous Scenario Generation, Financial Scenario Simulation.
    \item \textbf{Llama3-8B-Instruct}: Emotive Scenarios, Perspective Scenarios, Multilingual Scenarios.
    \item \textbf{GPT-3.5-turbo}: Historical Scenarios, Poetic Scenarios, Satirical Scenarios.
\end{itemize}
The distinct distribution of categories across these models highlights \model's ability to pinpoint model-specific weaknesses effectively.

\section{Iterative Improvement on Categories}
\label{appendix: iterative_improvement}
In Table \ref{tab:iter_improve_on_cat}, we present the top five ISR categories and the overall changes in ISR before and after iterative training on \texttt{llama2-13b-chat}. The ISR changes demonstrate a noticeable reduction in the frequency of similar types of errors, highlighting the effectiveness of the iterative training process.

\end{document}